\documentclass[11pt]{article}

\usepackage[final]{acl}

\usepackage{wrapfig}
\usepackage{amssymb}
\usepackage{algorithm}
\usepackage{algpseudocode}
\usepackage{multirow}
\usepackage{booktabs} 
\usepackage{subcaption}  
\usepackage{graphicx}     

\usepackage{tcolorbox}
\tcbuselibrary{breakable} 

\usepackage{times}
\usepackage{latexsym}
\usepackage{amsmath}
\usepackage{colortbl}  

\usepackage[T1]{fontenc}

\usepackage[utf8]{inputenc}

\usepackage{microtype}

\usepackage{inconsolata}

\usepackage{graphicx}

\title{WebAnchor: Anchoring Agent Planning to Stabilize Long-Horizon Web Reasoning}

\author{
\textbf{Xinmiao Yu}\thanks{Work done during the author's internship at Tongyi Lab.}, 
\textbf{Liwen Zhang}, 
\textbf{Xiaocheng Feng}, 
\textbf{Yong Jiang}~\footnotemark[2], \\
\textbf{Bing Qin}\thanks{Correspondence.}, 
\textbf{Pengjun Xie}, 
\textbf{Jingren Zhou} \\
Tongyi Lab, Alibaba Group
}

\begin{document}
\maketitle
\begin{abstract}

Large Language Model(LLM)-based agents have shown strong capabilities in web information seeking, with reinforcement learning (RL) becoming a key optimization paradigm. However, planning remains a bottleneck, as existing methods struggle with long-horizon strategies. 
Our analysis reveals a critical phenomenon—\textit{plan anchor}—where the first reasoning step disproportionately impacts downstream behavior in long-horizon web reasoning tasks. Current RL algorithms, fail to account for this by uniformly distributing rewards across the trajectory.
To address this, we propose Anchor-GRPO, a two-stage RL framework that decouples planning and execution. In Stage 1, the agent optimizes its first-step planning using fine-grained rubrics derived from self-play experiences and human calibration. In Stage 2, execution is aligned with the initial plan through sparse rewards, ensuring stable and efficient tool usage. We evaluate Anchor-GRPO on four benchmarks: BrowseComp, BrowseComp-Zh, GAIA, and XBench-DeepSearch. Across models from 3B to 30B, Anchor-GRPO outperforms baseline GRPO and First-step GRPO, improving task success and tool efficiency. Notably, WebAnchor-30B achieves 46.0\% pass@1 on BrowseComp and 76.4\% on GAIA. Anchor-GRPO also demonstrates strong scalability, getting higher accuracy as model size and context length increase.

\end{abstract}

\section{Introduction}



Reinforcement Learning (RL) have significantly enhanced the capabilities of autonomous, tool‑augmented agents built on large language models (LLMs) for web information seeking~\citep{jin2025search, li2025websailor, zhang2025landscapeagenticreinforcementlearning}. These agents often referred to as deep research~\citep{dr, groksearch, team2025tongyi}, go beyond static retrieval and instead iteratively formulate queries, invoke external tools, and collect evidence from diverse online sources to answer complex questions.

\begin{figure}[t]
    \centering
    \begin{subfigure}[b]{\linewidth}  
        \centering
        \includegraphics[width=\linewidth]{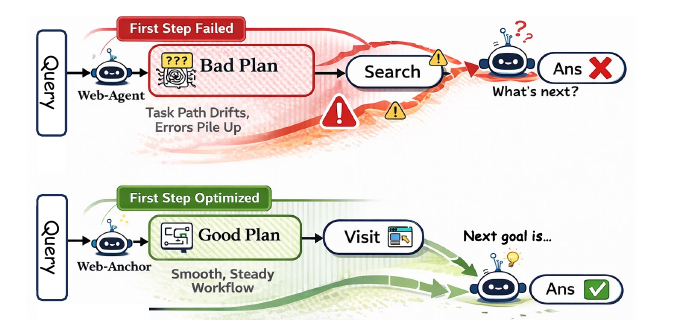} 
        \caption{The plan anchor phenomenon, where the first-step decision disproportionately affects the trajectory's success.}
        \label{fig:anchor}
    \end{subfigure}

    \begin{subfigure}[b]{0.48\linewidth}
        \centering
        \includegraphics[width=\linewidth]{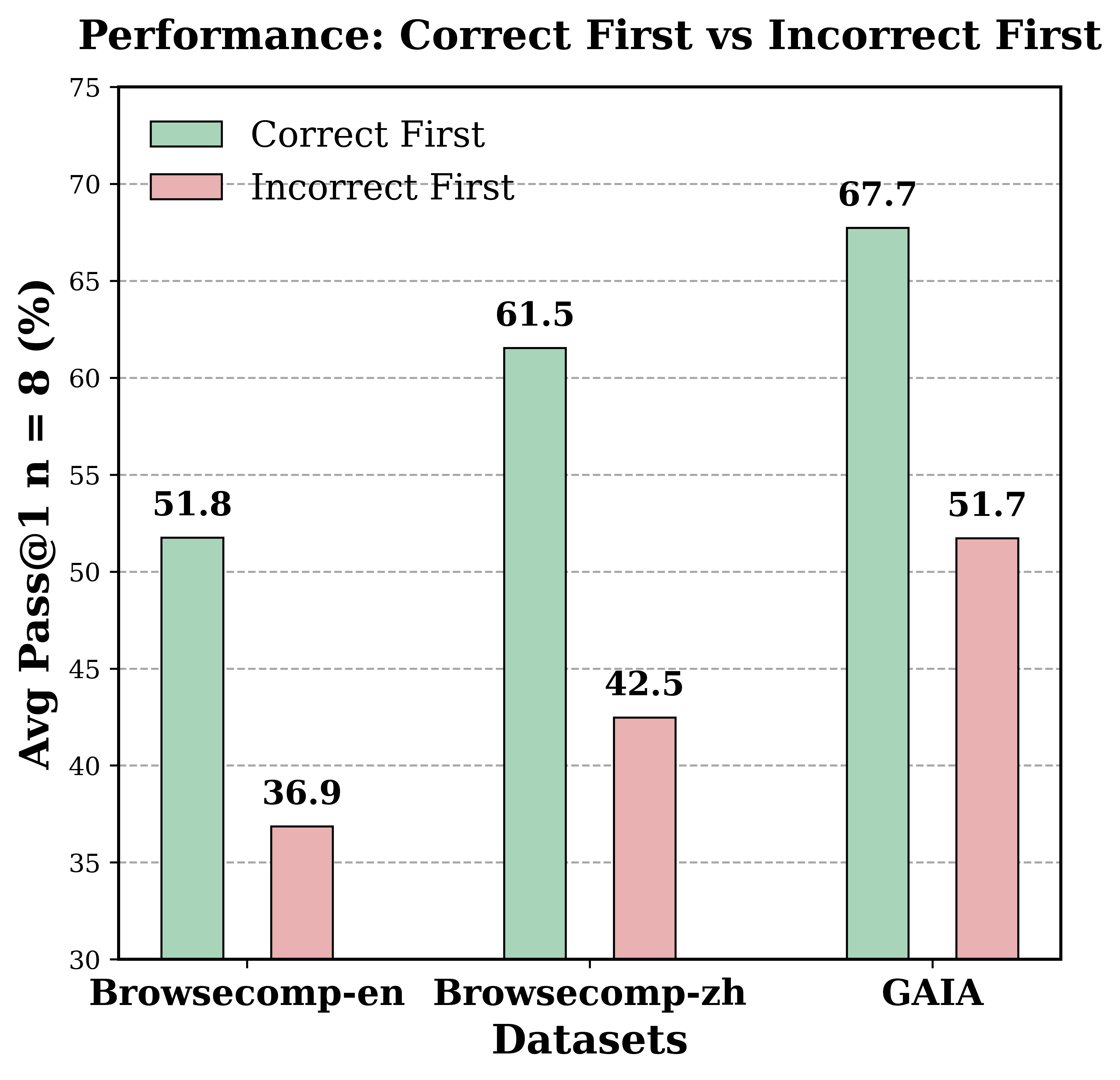}
        \caption{Impact of the first step on downstream task accuracy.}
        \label{fig:motivation1}
    \end{subfigure}
    \hfill
    \begin{subfigure}[b]{0.48\linewidth}
        \centering
        \includegraphics[width=\linewidth]{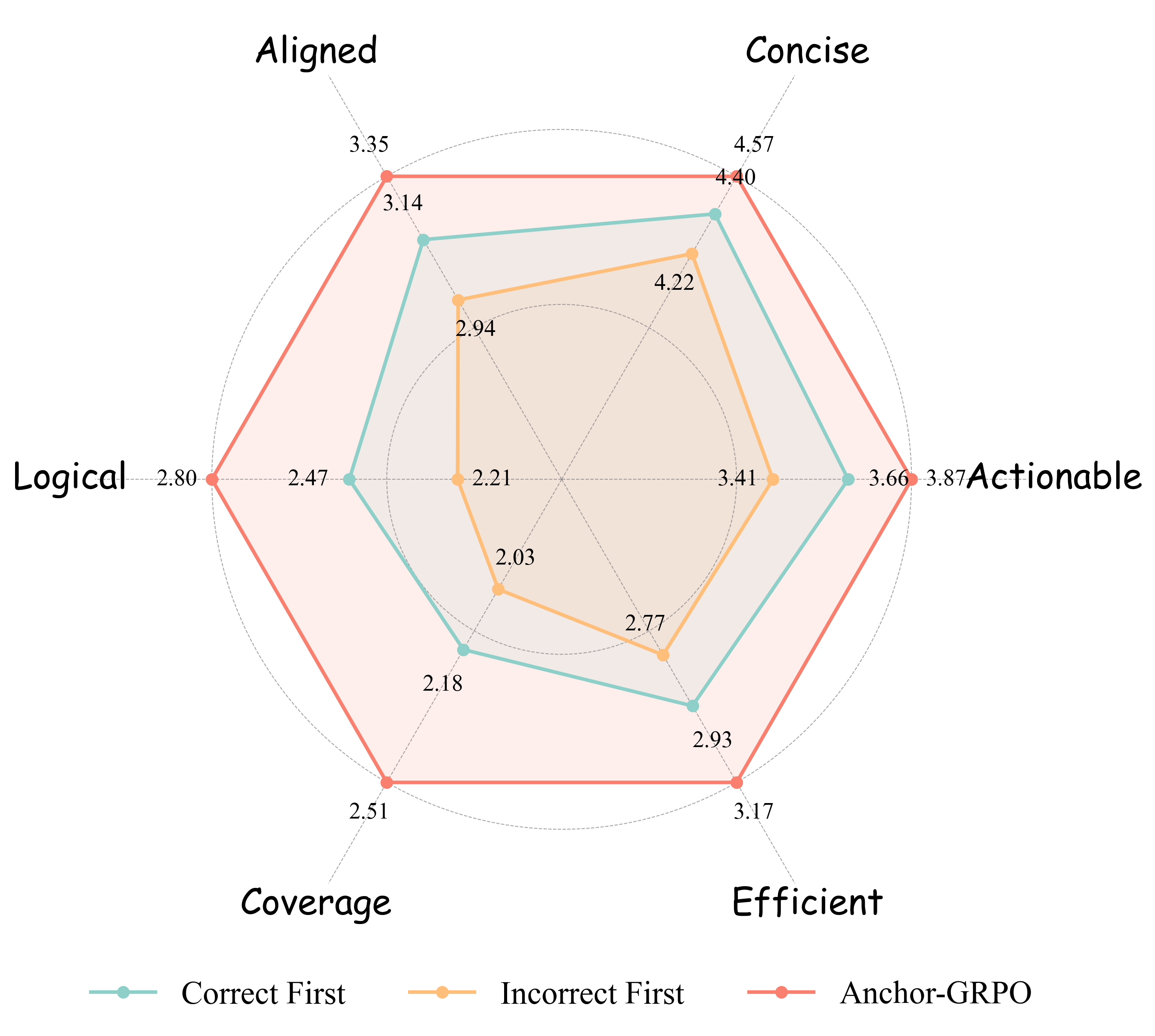}
        \caption{Plan rubrics visualized through a radar chart.}
        \label{fig:motivation_radar}
    \end{subfigure}

    \caption{Illustrating the plan anchor phenomenon, the critical role of the first step in task accuracy, and the use of plan rubrics to guide optimization.}
    \label{fig:abs_fig}
\end{figure}

Despite recent advancements, deep research agents still struggle with long-horizon planning and maintaining strategy coherence~\citep{erdogan2025plan, qiao2025webresearcher}. Increasing tool usage or managing context alone does not ensure consistent multi-step reasoning; instead, accumulated errors often lead to performance degradation~\citep{liu2025webexplorer}.
While existing reinforcement learning (RL) methods—such as context management, reward shaping, and entropy-based optimization—mitigate certain aspects, there are still gaps to address planning instability and the lack of a comprehensive long-term strategy~\citep{chung2025evaluating, wu2025resum, zhao2025repurposing, dong2025agenticreinforcedpolicyoptimization}.
A crucial insight is that the ability to maintain coherent long-horizon behavior hinges on early decisions~\citep{agentfounder2025}. Recent work highlights that not all reasoning steps are equally important~\citep{bogdan2025thought}. In particular, the first planning step often acts as a structural anchor—shaping exploration, tool usage, and evidence integration~\citep{dong2025emergent}. Initial missteps can trigger cascading failures and destabilize the entire trajectory~\citep{sui2025meta, sinha2025illusion}.
Motivated by this, we identify a critical phenomenon in long-horizon web agents, which we term the \textbf{\textit{plan anchor}}, as shown in Fig.~\ref{fig:anchor}. Our motivation experiments (Fig.~\ref{fig:motivation1}) demonstrate substantial performance drops in Avg Pass@1 across BC-ZH, BC-EN, and GAIA benchmarks due to incorrect first steps—by 28.7\%, 30.9\%, and 23.6\%, respectively. These results underscore the importance of designing reward schemes that explicitly prioritize the impact of the first step.

To address this, we introduce \textbf{\textit{Anchor-GRPO}}, a two-stage RL framework that separates planning and execution. The first stage focuses on optimizing the initial planning step. In this stage, we develop the \textbf{\textit{Plan Rubrics Learner}}, by analyzing both successful and failed experiences, the learner adaptively refines essential planning criteria, such as task decomposition, goal alignment, and tool selection. The final rubrics assign higher scores to correct plans, as shown in Figure~\ref{fig:motivation_radar}, confirming the learner’s ability to capture important planning capabilities. These refined rubrics are incorporated as reward signals in the RL process, guiding the agent to improve its planning capabilities. In Stage 2, the focus shifts to execution. Sparse rewards align execution with the initial plan, ensuring stability throughout the reasoning process and addressing long-horizon credit assignment challenges. By combining the power of plan rubrics and execution alignment, \textbf{\textit{Anchor-GRPO}} enables agents to \textit{plan first, then act}, resulting in a more reliable approach which can maintain consistency over long reasoning trajectories.

Our contributions are threefold:
\begin{itemize}
  \item \textbf{Plan Anchor Phenomenon in Long-Horizon Web Reasoning:} We identify the critical phenomenon of \textbf{\textit{plan anchor}}, where a first-step decision disproportionately impacts the success of the entire trajectory, highlighting the importance of the initial planning step in long-horizon web reasoning.
    
    \item \textbf{Experience-based Plan Rubrics Learner:} We propose the Plan Rubrics Learner framework, which adaptively learns key dimensions of effective long-horizon plans from self-play experiences. The learnt rubrics will guide the agent in optimizing the first planning step for improved reasoning and task success.
    
    \item \textbf{Anchor-GRPO and Its Superior Performance:} 
    We introduce Anchor-GRPO, a two-stage RL framework that enhances task success and long-horizon reasoning, outperforming existing methods on several challenging benchmarks. It also demonstrates scalability across model sizes and context lengths, making it suitable for more complex tasks.
\end{itemize}

    
    


\section{Preliminary}

\paragraph{Agentic Web Reasoning.}
We follow the standard ReAct-style agentic workflow \citep{yao2023react}, where an LLM-based agent interleaves \emph{Thought}, \emph{Action}, and \emph{Observation}.  
At step $t$, the agent reads the trajectory history $\mathcal{H}_{t-1}$ and produces a reasoning trace $\tau_t$ and an executable action $a_t$, after which the environment returns an observation $o_t$.  
A $T$-step rollout is defined as:
\[
\mathcal{H}_T = (q, \tau_1, a_1, o_1, \dots, \tau_T, a_T, o_T, \tau_{T+1}, a_{T+1}),
\]
where $q$ is the task query and $a_{T+1}$ denotes the final answer.  
At each step, the policy model samples:
\[
(\tau_t, a_t) \sim \pi_\theta(\cdot \mid \mathcal{H}_{t-1}).
\]

\paragraph{Tool Design.}
Following standard web agent designs~\citep{li2025websailor, gao2025turnsunlockinglonghorizonagentic}, we define the action space with two tools for web exploration:

\begin{itemize}
    \item \textbf{Search}: This tool issues top-$k$ web search queries to retrieve relevant snippets and URLs. It accepts natural language inputs and returns structured results from the Google Search API, including titles, snippets, and hyperlinks.

    \item \textbf{Visit}: Given a specific URL, the agent can browse the full page content and extract factual evidence. We use a language model browser to simulate document-level reading and structured content extraction.
\end{itemize}

Each action $a_t$ is thus instantiated as either \texttt{Search(query)} or \texttt{Visit(url)}. These tools enable multi-hop evidence gathering and decision-making under long-horizon uncertainty.

\section{Methodology}

In this section, we introduce a two-stage training framework Anchor-GRPO to optimize long-horizon reasoning in web agents. In the first stage, a \textbf{\textit{Plan Rubrics Learner}} is used to derive effective planning criteria from past experiences, which are used to optimize the agent’s initial planning with a dense reward. In the second stage, the agent’s execution is optimized based on sparse rewards, with both stages jointly trained to align planning and execution.

\subsection{Anchor-GRPO Overview}


Anchor-GRPO is a two-stage reinforcement learning framework that integrates planning and execution within a single policy model. In the first stage, the model acts as a \textit{Planner} (\( \pi_{\text{planner}} \)), generating an initial plan from the user query, optimized with dense rewards to improve task decomposition and planning quality. In the second stage, the model functions as an \textit{Executor} (\( \pi_{\text{executor}} \)), executing the plan via tool interactions and receiving sparse rewards based on task success.
Although both stages share the same model parameters, their training is decoupled through masked credit assignment: dense signals guide planning updates, while execution is refined using sparse feedback. This phased optimization aligns planning and execution without requiring separate models. We detail the reward design and masking-based credit assignment mechanism in the following sections.

\begin{figure}[htbp]
    \centering
    \includegraphics[width=1.0\linewidth]{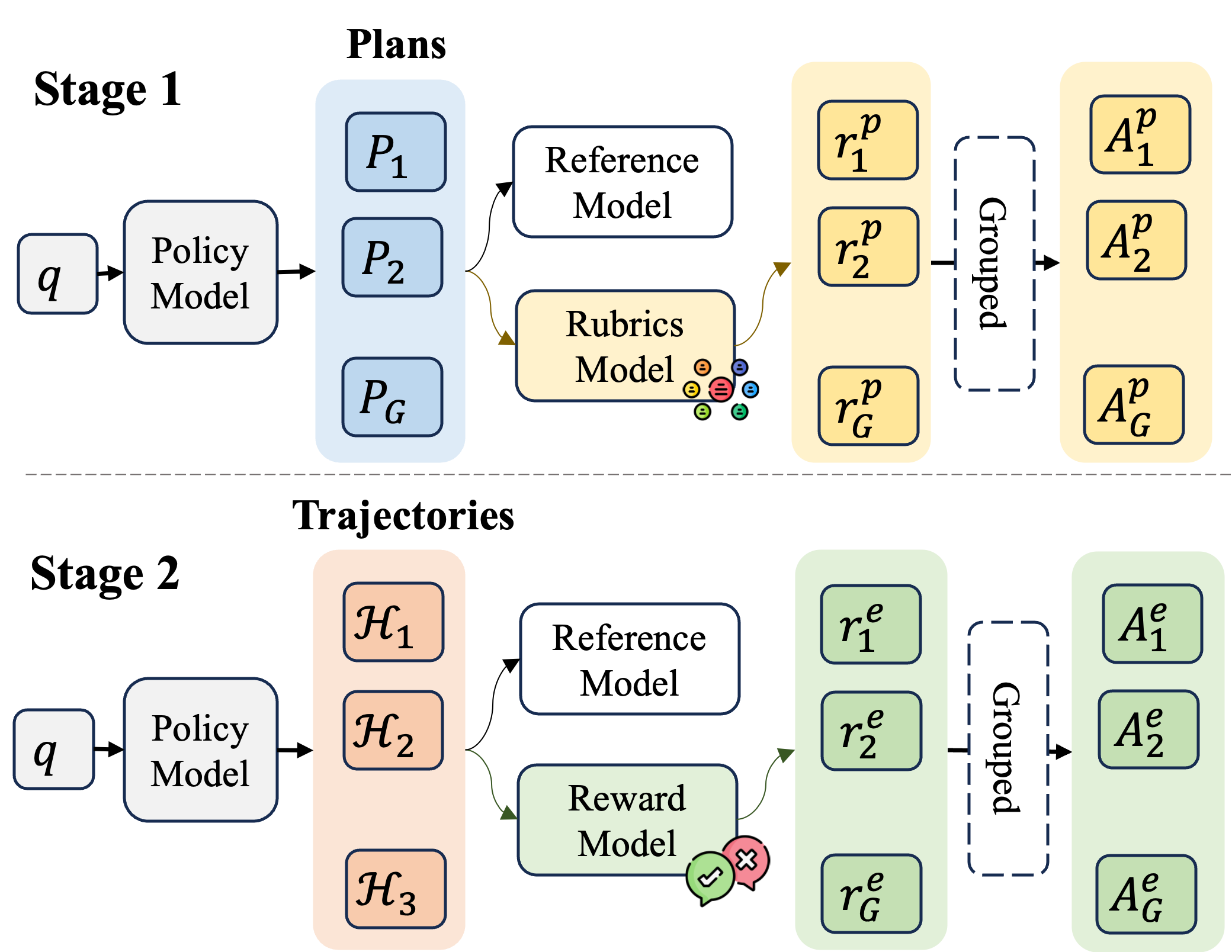} 
    \caption{Anchor-GRPO framework. Stage 1 optimizes the initial plan using the Rubrics Model, providing dense rewards. Stage 2 refines the trajectory with sparse rewards, ensuring alignment with the plan.}
    \label{fig:my_label}
\end{figure}

\begin{figure*}[htbp]
    \centering
    \includegraphics[width=1.0\linewidth]{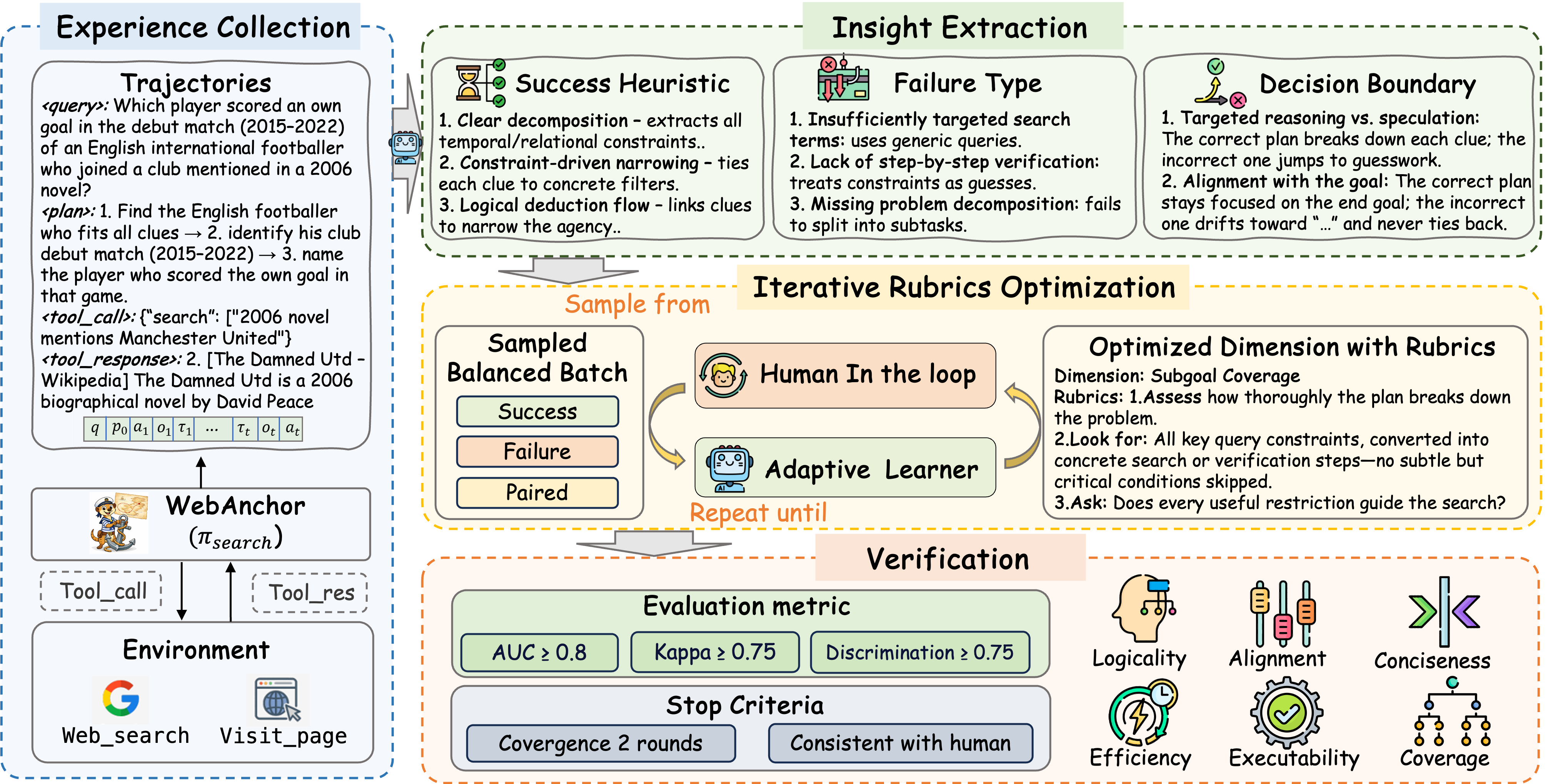} 
    \caption{Overview of the Plan Rubrics Learner and verification process, illustrating how WebAnchor collects experiences, extracts insights, iteratively optimizes rubrics, and verifies plan quality with human feedback.}
    \label{fig:main}
\end{figure*}



\subsection{Stage 1: Anchor Plan Optimization}

The first stage improves the agent’s initial planning by using Plan Rubrics learned from past experiences to define and reward good plans. This addresses the challenge of optimizing the first step in RL, ensuring each trajectory starts with a high-quality strategy and enhancing long-horizon performance.

\subsubsection{Plan Rubrics Learner}

The Plan Rubrics Learner (Figure~\ref{fig:main}) distills planning principles from a large corpus of agent trajectories collected during web-agent training and evaluation. Through iterative refinement which guided by an LLM and human-in-the-loop feedback, it learns structured rubrics across key dimensions, each with fine-grained criteria. The resulting rubrics align with human judgments and reliably distinguish between correct and incorrect plans, enabling effective reward shaping for downstream planning.

\paragraph{Insight Extraction}

We extract planning insights from a diverse set of task trajectories collected during prior agent interactions, spanning both successful executions and failure cases, using three LLM-based functions:  
$\mathcal{F}_\text{success}$ and $\mathcal{F}_\text{fail}$ analyze correct and incorrect trajectories to identify effective heuristics and failure modes, while $\mathcal{F}_\text{paired}$ compares paired trajectories to infer decision boundaries.  
Each function outputs a tuple $s_i = (q_i, p_i, \text{insight}_i)$, where $q_i$ is the task query, $p_i$ the initial plan, and $\text{insight}_i$ the derived principle.  
We get the set of insights \( S = \{ s_1, s_2, \dots, s_n \} \), defines a manifold of high-quality plans, separating successful from flawed strategies, and drives rubric refinement.

\paragraph{Rubrics Optimization}
We begin with an initial set of heuristic rubrics defined over $m$ planning dimensions $\{d_1, \dots, d_m\}$ (e.g., task decomposition, goal alignment, tool selection).  
These rubrics are iteratively refined through alternating LLM-driven updates and human feedback.

At each iteration $t$, we sample a balanced batch $\mathcal{B}_t = \{\mathcal{B}_{\text{success}}, \mathcal{B}_{\text{failure}}, \mathcal{B}_{\text{paired}}\}$ from the insight set $S$, and update the rubrics using an LLM-based updater $\mathcal{F}_{\text{Update}}$:
\[
\mathcal{R}_{t+1} = \mathcal{F}_{\text{Update}}(\mathcal{R}_t, \mathcal{B}_t),
\]
where $r_t$ denotes the rubrics at step $t$.

After each epoch, we evaluate the rubrics on two criteria: (1) alignment with human judgments, and (2) ability to discriminate between correct and incorrect plans via learned decision boundaries. When needed, human annotators refine ambiguous or erroneous rubric items to guide the next update cycle.  
The process continues until convergence criteria are met, yielding a robust rubric set that effectively shapes planning rewards.

\subsubsection{Stage 1: Anchor Plan Optimization with Plan Rubrics}

Stage 1 optimizes the \textit{Anchor Plan}, where the agent’s initial planning step decomposes the task, sets subtask goals, and selects tools. This stage uses a dense reward from the learned Plan Rubrics to improve first-step planning quality.

To focus credit assignment on the initial decision, we mask all actions after the first planning step during policy updates. Only the logits for the \textit{Anchor Plan} are updated, while subsequent steps receive zero gradient.

\paragraph{Plan Rubrics Reward Definition}
Let $p$ denote the generated Anchor Plan for a given query. The Plan Rubrics define a normalized scoring function $\mathcal{R}(p) \in [0,1]$, which evaluates $p$ across $m$ planning dimensions $\{d_1, \dots, d_m\}$:
\begin{equation}
\mathcal{R}^{\text{plan}} = \mathcal{R}( p ) 
= \frac{1}{Z} \sum_{j=1}^{m} \phi_j(p),
\end{equation}
where:
\begin{itemize}
    \item $\phi_j(p) = \text{Judge}_{\text{LLM}}(p, d_j) \in [0, s_j^{\max}]$ is the LLM-based score for dimension $d_j$,
    \item $s_j^{\max}$ is the maximum achievable score for $d_j$,
    \item $Z = \sum_{j=1}^{m} s_j^{\max}$ is a normalization constant.
\end{itemize}
This reward is used as the immediate return for the first planning step in Stage 1, providing a dense, interpretable signal that aligns with experience-based planning quality.

\paragraph{Objective Function for Plan Optimization}
We optimize the \textit{Anchor Plan} using a modified GRPO objective that operates only on initial plans. We sample a group of $G$ candidate plans $\{p_i\}_{i=1}^G$ from the old policy $\pi_{\theta_\text{old}}(\mathcal{P} \mid q)$ and maximize:

\begin{equation}
\begin{aligned}
\mathcal{J}_\text{plan}(\theta) &= \mathbb{E}_{(q, a) \sim \mathcal{D}, \{ p_i \}^G_{i=1} \sim \pi_{\theta_\text{old}}(\mathcal{P}|q)} \Bigg[ \frac{1}{G} \sum_{i=1}^{G} \\
& \quad \frac{1}{|p_i|}\sum_{j=1}^{|p_i|} \min \left( r_{i,j}(\theta) \hat{A}^{\text{plan}}_{i,j}, \right. \\
& \quad \left. \text{clip}(r_{i,j}(\theta), 1 - \epsilon_{\text{low}}, 1 + \epsilon_{\text{high}}) \hat{A}^{\text{plan}}_{i,j} \right) \Bigg],
\label{eq:plan_obj}
\end{aligned}
\end{equation}

where $r_{i,j}(\theta) = \frac{\pi_\theta(p_{i,j} \mid q, p_{i,<j})}{\pi_{\theta_\text{old}}(p_{i,j} \mid q, p_{i,<j})}$ is the importance sampling ratio at token $j$ of plan $i$, $\hat{A}^{\text{plan}}_{i,j}$ is the advantage estimate derived from the rubric $\mathcal{R}^{\text{plan}}_i$ via group normalization, and $\epsilon_{\text{low}}$, $\epsilon_{\text{high}}$ control the clipping range to stabilize policy updates.

\paragraph{LLM Masking for First-Step Update} In this approach, the update for the planning policy \( \pi_{\text{plan}} \) is confined to the first step of the trajectory. The LLM's outputs for subsequent steps are masked, ensuring that only the first step's plan influences the update. This isolates the planner's optimization to the initial decision, preventing interference from later trajectory interactions or the executor policy.

\subsection{Stage 2: Trajectory Level Executor Optimization}

\paragraph{Executor Reward}
In the second stage, the executor is trained to follow the optimized plan using a sparse task-completion reward. The reward is assigned only at the end of the episode based on whether the final answer exactly matches the ground truth:
\begin{equation}
\mathcal{R}^{\text{exec}} =
\begin{cases}
1 & \text{if } \texttt{ExactMatch}(\text{Answer}, \text{GT}), \\
0 & \text{otherwise}.
\end{cases}
\end{equation}
All intermediate steps receive zero reward (\( r^{\text{exec}}_t = 0 \) for \( t < T \)), with the full reward \( r^{\text{exec}} \) assigned only at termination (\( t = T \)), encouraging the executor to follow the plan and achieve task success.




\paragraph{Objective for Execution}
In Stage 2, we optimize the executor policy at the trajectory level. We sample a group of $G$ rollouts $\{ \mathcal{H}^{(i)} \}_{i=1}^G$ from the old policy $\pi_{\theta_\text{old}}(\mathcal{H} \mid q)$ and maximize:

\begin{equation}
\begin{aligned}
\mathcal{J}_\text{exec}(\theta) &= \mathbb{E}_{(q, a) \sim \mathcal{D}, \{H^{(i)}\}^G_{i=1} \sim \pi_{\theta_\text{old}}(\mathcal{H}|q)} \Bigg[ \frac{1}{G} \sum_{i=1}^{G} \\
& \quad \frac{1}{|\mathcal{H}^{(i)}|} \sum_{j=1}^{|\mathcal{H}^{(i)}|} \min \left( r_{i,j}(\theta) \hat{A}^{\text{exec}}_{i,j}, \right. \\
& \quad \left. \text{clip}(r_{i,j}(\theta), 1 - \epsilon_{\text{low}}, 1 + \epsilon_{\text{high}}) \hat{A}^{\text{exec}}_{i,j} \right) \Bigg],
\end{aligned}
\end{equation}

where $r_{i,j}(\theta)$, $\epsilon_{\text{low}}$, and $\epsilon_{\text{high}}$ are defined identically to those in Equation~\eqref{eq:plan_obj}, while $\hat{A}^{\text{exec}}_{i,j}$ is the advantage estimate derived from the trajectory-level execution reward $\mathcal{R}^{\text{exec}}$ via group normalization.

\subsection{Plan Rubrics Evaluation and Validation}
Rubrics are validated at the end of each optimization cycle against manual plan outcomes using two metrics: AUC (measuring ranking quality) and Cohen's $\kappa$ (measuring agreement). We require
\[
\text{AUC} \geq 0.8, \quad \kappa \geq 0.75.
\]
If either threshold is not met, the rubrics will be revised. This iterative validation--revision loop continues until convergence, ensuring the rubrics remain aligned with task success criteria.

\definecolor{lightblue}{rgb}{0.9, 0.95, 1}  

\begin{table*}[t]
\centering
\small
\setlength{\tabcolsep}{4pt}
\begin{tabular}{lccccccccc}
\toprule

\multirow{3}{*}{Model} & \multirow{3}{*}{RL Algo}  & \multicolumn{2}{c}{GAIA} & \multicolumn{2}{c}{BrowseComp} & \multicolumn{2}{c}{BrowseComp-ZH} & \multicolumn{2}{c}{Xbench} \\
\cmidrule(lr){3-4} \cmidrule(lr){5-6} \cmidrule(lr){7-8} \cmidrule(lr){9-10}
& & Pass@1 & Pass@3 & Pass@1 & Pass@3 & Pass@1 & Pass@3 & Pass@1 & Pass@3 \\
\midrule
\multicolumn{9}{l}{\textit{Advanced Models}} \\
OpenAI-o3$^\dagger$ & - & \underline{70.5} & - & \underline{50.9} & - & \underline{58.1} & - & 66.7 & - \\
Claude-4-Sonnet$^\dagger$ & - & 68.3 & - & 12.2 & -& 29.1 & - & 64.6 & -  \\
Kimi-K2$^\dagger$ & - & 57.7 & - & 14.1 & - & 28.8 & - & 50.0 & - \\
DeepSeek-V3.1$^\dagger$ & - & 63.1 & - & 30.0 & -  & 49.2 & - & \underline{71.0} & -\\
\midrule
\multicolumn{9}{l}{\textit{Open-source Agents $\leq$ 32B}} \\
R1-Searcher-7B & - & 20.4 & - & 0.4 & - & 0.6 & - & 4.0 & - \\
WebThinker-RL & - & 48.5& - & 2.8 & - & 7.3 & - & 24.0 & - \\
WebDancer-QwQ & - & 51.5 & - & 3.8 & - & 18.0 & - & 39.0 & - \\
WebSailor-7B & - & 37.9 & - & 6.7 & - & 14.2 & - & 34.3 & - \\
WebSailor-32B & - & \underline{53.2} & -  & \underline{10.5} & - & \underline{25.5} & - & \underline{53.3} & - \\
\midrule

\multirow{3}{*}{\textbf{WebAnchor-3B}} 

& GRPO & 27.2 & \underline{43.6} & 5.1 & 8.7 & 12.3 & 19.2 & 31.6 & 55.0 \\
& First-step GRPO & \textbf{30.1} & \textbf{44.6} & \underline{6.9} & \underline{10.2} & \underline{12.9} & \underline{21.5} & \underline{34.7} & \underline{55.0} \\

\rowcolor{lightblue} &   \textbf{Anchor-GRPO}  & \underline{28.3} & 43.5 & \textbf{7.1} & \textbf{11.5} & \textbf{13.1} & \textbf{25.0} & \textbf{37.7} & \textbf{56.0}  \\

\midrule

\multirow{3}{*}{\textbf{WebAnchor-7B}} 
& GRPO & 33.0 & 44.7 & 6.3 & 11.7 & \underline{17.5} & \underline{31.5} & \underline{40.7} & \underline{56.0} \\
& First-step GRPO & \textbf{38.5} & \textbf{55.1} & \underline{9.3} & \underline{16.3} & 16.4 & 31.2 & 37.7 & 52.0\\

\rowcolor{lightblue} &  \textbf{Anchor-GRPO} & \underline{37.8} & \underline{51.5} & \textbf{11.6} & \textbf{20.7} & \textbf{23.8} & \textbf{36.7} &  \textbf{45.0} & \textbf{66.0} \\

\midrule

\multirow{3}{*}{\textbf{WebAnchor-30B}} 
& GRPO & \underline{75.8} & \textbf{87.6} & \underline{44.0} & \underline{65.0} & 46.2 &61.4 &71.0 & 84.0 \\
& First-step GRPO & 74.8 & \underline{87.4} & 43.3 & 66.0 & \underline{47.9} & \underline{63.2}  & \underline{72.0} & \textbf{88.0} \\
\rowcolor{lightblue} &  \textbf{Anchor-GRPO} & \textbf{76.4} & 86.4 & \textbf{46.0} & \textbf{67.5} & \textbf{48.8} & \textbf{65.1} & \textbf{75.1} & \underline{85.0} \\

\bottomrule
\end{tabular}
\caption{Performance comparison of the proposed Anchor-GRPO method across different model sizes and RL algorithms, evaluated on multiple benchmarks (GAIA, BrowseComp, BrowseComp-ZH, Xbench).}
\label{tab:anchor_grpo_main}
\end{table*}

\section{Experiments}
In this section, we present a series of experiments designed to evaluate the effectiveness and scalability of Anchor-GRPO, addressing three critical questions:

\paragraph{1. Performance of Anchor-GRPO:}
Does WebAnchor outperform strong external agents, including proprietary models (e.g., OpenAI, DeepSeek-V3.1) and open-source models (e.g., R1-Searcher~\citep{song2025r1}, WebSailor), as well as internal baselines such as GRPO and First-step GRPO, across challenging agentic benchmarks?

\paragraph{2. Plan Quality and Downstream Impact:}
How does Anchor-GRPO enhance the quality of initial plans, particularly in terms of Subgoal Coverage, Goal Alignment, and Tool Efficiency? More importantly, how does this improvement in planning quality translate to better downstream execution and task success?

\paragraph{3. Scalability Potential:}
Is the two-stage design of Anchor-GRPO well-suited for future scaling? We examine whether performance consistently improves as model sizes increase (ranging from 3B to 30B) and context lengths grow (from 16k to 64k), suggesting strong potential for continued improvement as model capacity and complexity advance.

\paragraph{Ablation Studies:}
We conduct ablation studies that evaluate the necessity of core elements—such as two-stage optimization, rubric-based reward shaping, and first-step planning. Additionally, we analyze how the quality of the Anchor Plan impacts tool usage efficiency and overall task performance.

\subsection{Experimental Setup}

\subsubsection{Benchmarks and Metrics}

\paragraph{Benchmarks} We evaluate our method on four challenging benchmarks for web-based information-seeking tasks: BrowseComp\_en \citep{bc_en}, BrowseComp\_zh \citep{bc_zh}, XBench-DeepSearch \citep{xbench}, and GAIA \citep{mialon2023gaia}. 

\paragraph{Metrics} We evaluate using Pass@1 and Pass@3, which measure the success rates of finding the correct answer in the first and top-three rollouts, respectively. Pass@1 is averaged over three runs for stability. We use Qwen-2.5-72B as scoring model.

\subsubsection{Baselines}
We conduct experiments on models ranging from 3B to 30B: including WebSailor-3B~\footnote{~\url{https://huggingface.co/Alibaba-NLP/WebSailor-3B}}
, WebSailor-7B~\footnote{~\url{https://huggingface.co/Alibaba-NLP/WebSailor-7B}}
, and Tongyi-DR-30B~\footnote{~\url{https://huggingface.co/Alibaba-NLP/Tongyi-DeepResearch-30B-A3B}}
, which are fine-tuned on synthetic information-seeking data for multi-turn tool use and serve as the base policies. We compare three RL settings, including our method: \textbf{1. GRPO}, which applies GRPO over the entire trajectory using exact match reward for all planning and execution steps; and \textbf{2. First-step GRPO}, a two-stage approach that first optimizes the initial planning step and then optimizes the entire trajectory. Both stages use sparse (0/1) rewards based on final answer exact match. Our method, \textbf{3. Anchor-GRPO}, also uses a two-stage process but leverages Plan Rubrics–derived dense rewards for initial plan optimization, followed by sparse task-completion rewards during execution.


\paragraph{Training Setup} We use 1,000 high-quality examples from an in-house wiki corpus, filtered by task difficulty. The training process takes place in a virtual wiki environment, replaces the real web to ensure faster request speeds and improved stability. Our experiments are conducted on 64 GPUs, with batch size of 32 and rollout num of 8.

\begin{figure*}[t]
    \centering
    \begin{subfigure}[b]{0.32\textwidth}
        \centering
        \includegraphics[width=\linewidth]{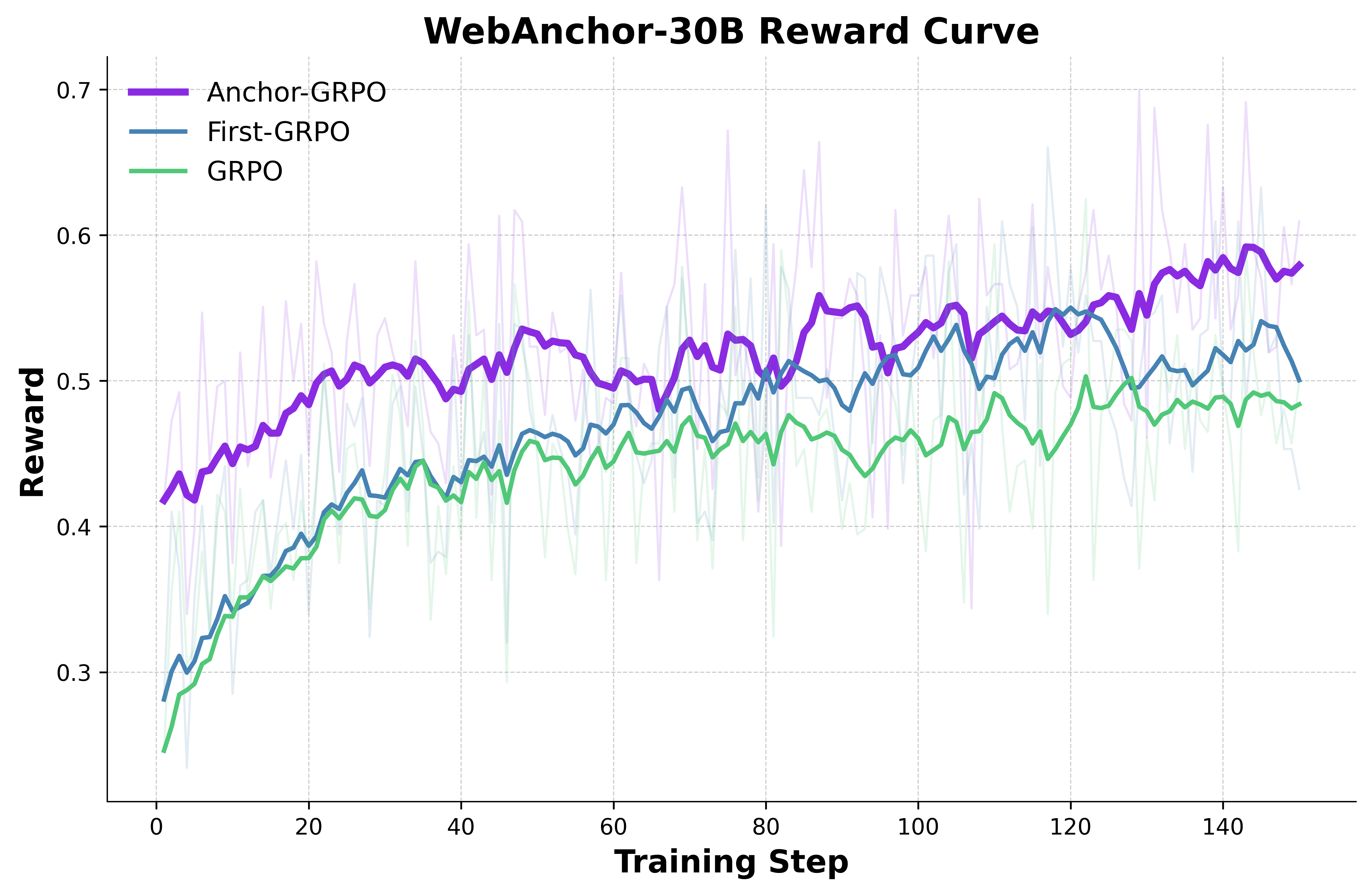}
        \caption{Training dynamics comparison across GRPO, First GRPO, and Anchor-GRPO.}
        \label{fig:sub1}
    \end{subfigure}
    \hfill
    \begin{subfigure}[b]{0.32\textwidth}
        \centering
        \includegraphics[width=\linewidth]{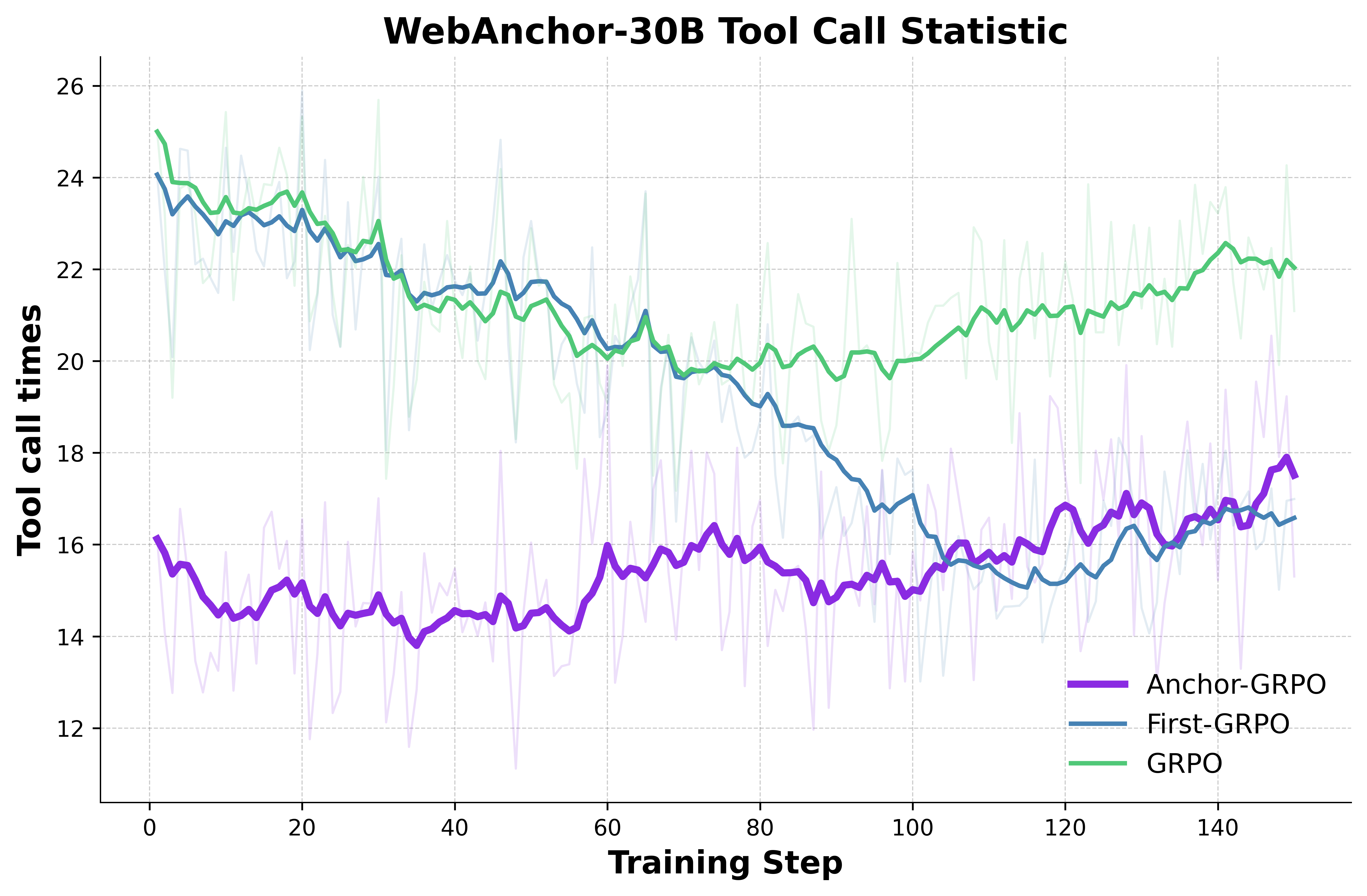}
        \caption{Tool-call usage comparison across GRPO, First GRPO, and Anchor-GRPO.}
        \label{fig:sub2}
    \end{subfigure}
    \hfill
    \begin{subfigure}[b]{0.32\textwidth}
        \centering
        \includegraphics[width=\linewidth]{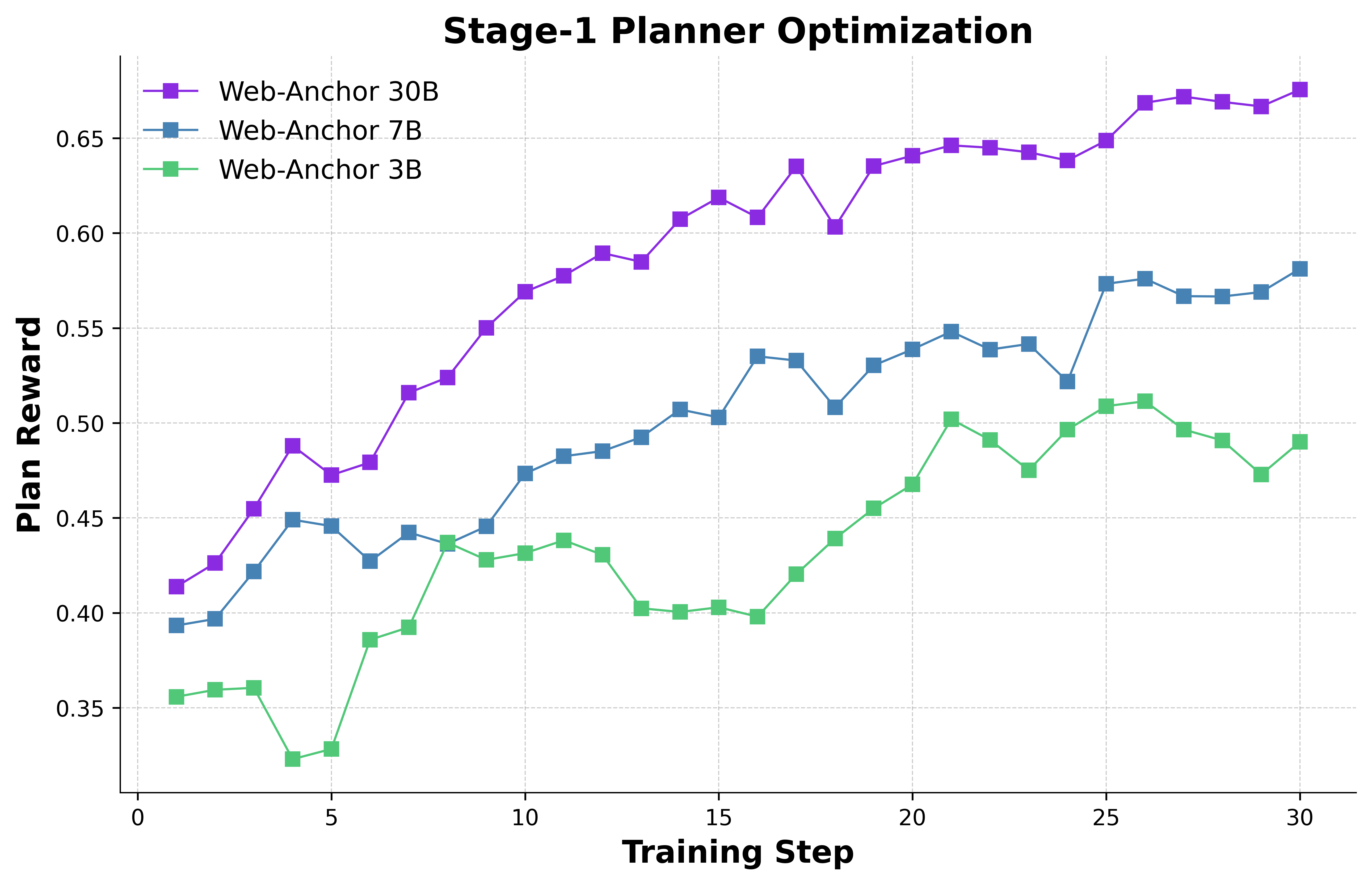}
        \caption{Rubrics reward scaling of Anchor-GRPO across model sizes (3B–30B).}
        \label{fig:sub3}
    \end{subfigure}
    \caption{Performance and behavior analysis of Anchor-GRPO versus baselines, including training convergence, tool efficiency, and reward robustness across model scales.}
    \label{fig:three-subfigs}
\end{figure*}

\subsection{Main Results}
\paragraph{Anchor-GRPO Consistently Outperforms Baselines and Achieves SOTA Performance Across Different Model Sizes}
Anchor-GRPO outperforms both baseline GRPO and other models, achieving SOTA performance in Pass@1 and Pass@3 across multiple benchmarks. For example, WebAnchor-30B achieves \textbf{46.0\% Pass@1} on BrowseComp, surpassing baseline GRPO (42.0\%) and First-step GRPO (41.3\%). In GAIA, WebAnchor-30B achieves \textbf{76.4\% Pass@1}, outperforming WebSailor-32B (53.2\%) and OpenAI-o3 (70.5\%). These results highlight Anchor-GRPO's ability to improve task completion rates and surpass existing methods in long-horizon reasoning.

\paragraph{First-GRPO Demonstrates the Effectiveness of Two-Stage Training, While Anchor-GRPO Further Optimizes with Plan Rubrics Reward}
First-step GRPO demonstrates the benefits of two-stage training by optimizing the first step independently. Anchor-GRPO further enhances performance by integrating a \textbf{Plan Rubrics Reward} in Stage 1. This reward significantly improves the first-step planning, as shown by the \textbf{2.7\% improvement in Pass@1} for WebAnchor-30B over First-step GRPO at BC, underscoring the advantage of planning optimization through structured rubrics.

\paragraph{Anchor-GRPO Exhibits Strong Scalability Across Different Model Sizes}
Anchor-GRPO shows excellent scalability, with performance improving from 3B to 30B models. WebAnchor-30B achieves \textbf{76.4\% Pass@1} on GAIA, significantly outpacing WebAnchor-7B (37.8\% Pass@1). In XBench, WebAnchor-30B reaches \textbf{75.1\% Pass@1}, showing robust performance as model size increase, proving its potential for complex tasks.

\subsection{Training Dynamics}

\paragraph{Reward Dynamics}
The WebAnchor-30B training curve shows in ~\ref{fig:sub1} that Anchor-GRPO consistently outperforms both First-GRPO and GRPO, with a steady increase in rewards. This improvement is driven by Stage 1’s Plan optimization, which strengthens the agent’s first-step planning. By decoupling planning and execution, Anchor-GRPO provides a stable foundation for higher performance throughout training.

\paragraph{Tool Calling Dynamics}
As shown in ~\ref{fig:sub2}, Anchor-GRPO exhibits more efficient and stable tool usage compared to both GRPO and First-GRPO. While First-GRPO reduces tool calls, it limits task success by restricting exploration. In contrast, Anchor-GRPO optimizes tool usage, balancing efficiency with performance, resulting in better task completion.

\subsection{Ablation Studies}
\begin{table*}[t]
\centering
\small
\setlength{\tabcolsep}{4pt} 
\begin{tabular}{lcccccc}
\toprule
\textbf{Ablation Settings} & \textbf{Pass@1} & \textbf{Pass@3} & \textbf{Coverage} & \textbf{Alignment} & \textbf{Efficiency} & \textbf{Avg. Tool Calls} \\
\midrule
\multicolumn{7}{l}{\textit{Two-Stage Training Strategy}} \\
\quad Full GRPO (baseline)         & \underline{44.0} & \underline{65.0} & 36.0 & 61.2 & 53.8 & 15.5 \\
\quad Only Planner Updated (Stage-1) & 42.0 & 57.0 & \underline{48.2} & \textbf{67.2} & \textbf{65.4} & 14.9 \\
\rowcolor{lightblue} \quad  \textbf{Two-Stage (Planner + Executor)} & \textbf{46.0} & \textbf{67.5} & \textbf{50.2}& \underline{67.0} & \underline{63.4}  & 17.3 \\
\midrule
\multicolumn{7}{l}{\textit{First-Step vs. Other-Step Update}} \\
\quad Random Step GRPO             & 33.0 & 38.5 & \underline{39.0} & \underline{43.6} & \underline{46.6} & 16.2 \\
\quad Last Step GRPO               & \underline{36.1} & \underline{40.8} & 33.4 & 41.4 & 42.8 & 13.0 \\
\rowcolor{lightblue} \quad  \textbf{First-Step GRPO} & \textbf{43.3} & \textbf{66.0} & \textbf{44.4} & \textbf{58.2} & \textbf{56.6} & 14.6 \\
\midrule
\multicolumn{7}{l}{\textit{Reward Design}} \\
\quad 0-1 Terminal Reward          & 43.3 & \underline{66.0} & 39.8 & 43.4 & 47.8 & 15.7 \\
\quad Naive Plan Reward            & \underline{44.2} & 63.4 & \underline{44.6} & \underline{52.2} & \underline{51.4} & 19.2 \\
\rowcolor{lightblue} \quad  \textbf{Planner Dense Reward} & \textbf{46.0} & \textbf{67.5} & \textbf{50.2}& \textbf{67.0} & \textbf{63.4} & 17.3 \\
\bottomrule
\end{tabular}
\caption{Ablation studies on Browsecomp\_en across three design dimensions. We compare how different settings affect performance, plan quality and tool call efficiency.}
\end{table*}

\paragraph{Two-Stage Training Strategy}  
We compare three settings: (i) standard GRPO (baseline), (ii) Stage-1-only, and (iii) full two-stage GRPO. While Stage-1-only yields the highest Alignment (67.2) and Efficiency (65.4), its Pass@1 (42.0) lags behind the full method. The two-stage approach achieves the best task success (Pass@1: \textbf{46.0}), demonstrating that joint optimization enables the agent to adaptively refine execution based on high-quality plans.

\paragraph{First-Step vs. Other-Step Update}  
We ablate which step is updated during GRPO: first, last, or a random intermediate step. First-Step GRPO achieves the highest Pass@1 (43.3) and Alignment (58.2), significantly outperforming Random Step (33.0) and Last Step (36.1). This confirms that optimizing the initial planning decision provides a critical anchor for long-horizon task completion.

\paragraph{Reward Design for Planning}  
We evaluate three reward schemes for the planner: (i) sparse 0–1 terminal reward, (ii) Naive Plan Reward, and (iii) our dense rubric-based reward. The dense rubric reward leads to the strongest performance (Pass@1: \textbf{46.0}), substantially improving over Naive (44.2) and terminal (43.3) rewards, highlighting the value of structured, multi-dimensional feedback in shaping effective plans.





\subsection{Scaling of Anchor-GRPO}

\begin{figure}[htbp]
    \centering
    \includegraphics[width=0.8\linewidth]{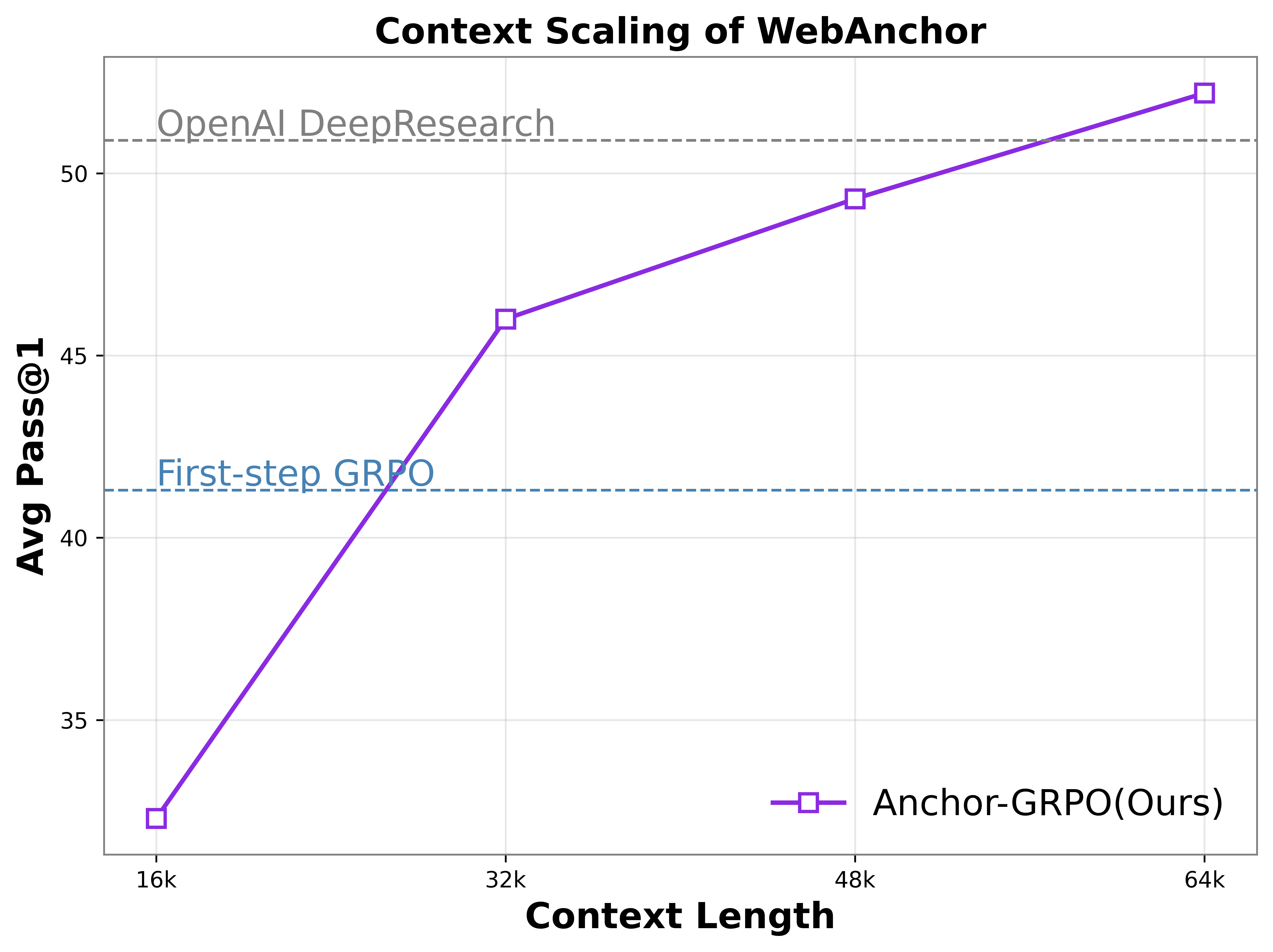} 
    \caption{Scaling of Anchor-GRPO}
    \label{fig:context_scale}
\end{figure}

\paragraph{Context length scaling}  
We evaluate Anchor-GRPO on BrowseComp\_EN with context lengths of 32k, 48k, and 64k. Results shows in ~\ref{fig:context_scale} consistent performance gains as context length increases, demonstrating effective scaling. This suggests that Anchor-GRPO will further benefit from models with larger context windows or greater capacity.

\paragraph{Plan rubrics reward scaling}  
As shown in ~\ref{fig:sub3}, WebAnchor models (3B, 7B, and 30B) all successfully converge on the plan rubrics reward, with reward values increasing monotonically with model size. This demonstrates strong scaling behavior of the learned rubrics with respect to model capacity.

\section{Related Work}
\paragraph{Long Horizon Web Reasoning}
Recent Deep Research (DR) agents tackle long-horizon web reasoning through multi-step planning, iterative refinement, and web-scale evidence synthesis. Moving beyond standard RAG, works like WebThinker~\citep{Li2025webthinker} and WebResearcher~\citep{qiao2025webresearcher} integrate Large Reasoning Models (LRMs) for deep thinking and self-correction. To manage unstructured evidence, structural frameworks such as WebWeaver~\citep{li2025webweaverstructuringwebscaleevidence} employ adaptive hierarchies to preserve coherence. These advances are benchmarked by DeepResearch Bench~\citep{du2025deepresearch} and extended to multimodal settings~\citep{geng2025webwatcher}), forming a robust foundation for autonomous scientific research.

\paragraph{Agentic Reinforcement Learning}

Agentic RL has emerged as a key paradigm, transforming LLMs from passive sequence generators to autonomous agents capable of environmental interaction and multi-step decision-making. Foundational surveys have formalized this transition, highlighting how RL optimizes both internal reasoning and external actions of agents~\citep{zhang2025landscapeagenticreinforcementlearning}. Multi-turn training frameworks like Agent-R1~\citep{cheng2025agentr1trainingpowerfulllm} and AgentGym-RL~\citep{xi2025agentgymrltrainingllmagents} enhance long-horizon performance and tool-use capabilities. Researchers are also addressing challenges like sparse feedback and robust tool integration through novel reward structures, as seen in VerlTool~\citep{jiang2025verltoolholisticagenticreinforcement}. These advancements, supported by meta-thinking frameworks such as ReMA~\citep{wan2025remalearningmetathinkllms} and verifiable reasoning models like RLVMR~\citep{zhang2025rlvmrreinforcementlearningverifiable}, create a comprehensive ecosystem where RL powers tool-augmented AI systems.

\section{Conclusion}
We present Anchor-GRPO, a two-stage reinforcement learning framework that decouples planning and execution to address the unique challenges of long-horizon web reasoning. By introducing Plan Rubrics Learner, structured criteria distilled from agent experiences, we enable dense and interpretable reward shaping that significantly improves plan quality. Our ablation studies confirm that (1) optimizing the first planning step acts as a critical anchor for downstream success, (2) joint planner-executor training yields superior task accuracy over planner-only optimization, and (3) rubric-based dense rewards are essential for effective policy learning. Evaluated on complex web research tasks, Anchor-GRPO achieves state-of-the-art performance, demonstrating that principled planning grounded in explicit reasoning standards is key to building stable agents.

\section*{Limitations}
In this work, we have focused on applying the method to web agents and related tasks. However, we believe the plan anchor phenomenon may also be relevant in other domains, and we look forward to exploring the potential of this method in those areas. WebAnchor still has significant room for improvement in the proposed plan rubrics, which could potentially lead to further performance gains.
We also hope that future research will continue to highlight the importance of first-step planning and introduce new optimization techniques to enhance its effectiveness.
\section*{Acknowledgments}

\bibliography{custom}
\newpage
\appendix
\section{Appendix}
\label{sec:appendix}

\subsection{Motivation Experiment Details}

We use the Tongyi-deepresearch-A30 model to generate three rounds of rollouts for each dataset: Browsecomp-en, Browsecomp-zh, and GAIA. We select queries that are neither all wrong nor all correct. We separately select the correct first step and the incorrect first step. Then, we fix the first step and generate 8 rollouts. We calculate the average Pass@8 for the correct first step and the incorrect first step. We found significant dropouts in BC-ZH, BC-EN, and GAIA, with drops of 28.76\%, 30.89\%, and 23.63\%, respectively. These results highlight the significant effect of the first step anchoring.

\subsection{Detailed Prompts}

\subsubsection{Insight Extraction Prompt}
\paragraph{Single Insight Extraction Prompt} This prompt is used to extract insight from single successful or failed trajectory.
\begin{tcolorbox}[colback=gray!15, colframe=gray!50, boxrule=0.5pt]
You are an expert in evaluating reasoning plans.

Query:
{question}

Plan (known to be {correctness}):
{plan}

Task:
Analyze what makes this plan {correctness}. Focus on:
- Specific behaviors that contribute to success (if correct)
- Critical flaws that cause failure (if incorrect)
- Concrete dimensions of planning quality

Output only a concise bullet-point list of insights.

\end{tcolorbox}

\paragraph{Paired Insight Extraction Prompt} This prompt is used to extract insight from single successful or failed trajectory.
\begin{tcolorbox}[colback=gray!15, colframe=gray!50, boxrule=0.5pt]
You are an expert in evaluating reasoning plans for web-based information-seeking tasks.

Below is a query and two plans: one that leads to the correct answer, and one that fails.

Query:
{question}

Correct Plan:
{correct\_plan}

Incorrect Plan:
{incorrect\_plan}

Task:
Analyze the key differences between these two plans. Specifically:
1. What does the correct plan do well that the incorrect plan misses?
2. Which dimensions of planning quality are most discriminative here?
3. Avoid vague statements; be concrete and grounded in the text.

Output only a concise bullet-point list of insights (no intro/outro)

\end{tcolorbox}
\subsubsection{Plan Rubrics Prompt}

\begin{tcolorbox}[
  breakable,
  width=\columnwidth,
  colback=gray!15,
  colframe=gray!50,
  boxrule=0.5pt,
  left=1mm,
  right=1mm,
  top=1mm,
  bottom=1mm
]
You are tasked with evaluating the following plan for a web information seeking task. Please score the plan on the following dimensions using a scale from 0 to 5:

- **0 = Very Poor**: Plan barely addresses the criteria or is largely incorrect.  
- **1 = Poor**: Plan meets only a small part of the criteria and is mostly ineffective.  
- **2 = Fair**: Plan is somewhat effective and meets basic aspects, though with flaws or inefficiencies.  
- **3 = Good**: Plan is reasonably effective and meets many of the criteria.  
- **4 = Very Good**: Plan is mostly effective and meets most criteria with minor issues.  
- **5 = Excellent**: Plan is highly effective, efficient, and well-structured.

Plan and Query:

- **Query**: <query>
- **Plan**: <plan>

Please rate the plan on the following dimensions, considering the **query** as the task goal and evaluating how well the plan addresses the task:

1. Goal Alignment
What to assess: Whether the plan stays focused on what the user actually wants.
Look for: A clear sense of the final answer—what kind of thing it is (a name, a number, a date) and how the query’s details shape it. The best plans show how each clue helps zero in on that target.
Ask yourself: Does this plan really understand what needs to be found—and why?

2. Subgoal Coverage
What to assess: How thoroughly the plan breaks down the problem.
Look for: All key constraints from the query (time, place, people, numbers, relationships) turned into concrete search or verification steps. Strong plans don’t skip subtle but critical conditions.
Ask yourself: Has every meaningful restriction been accounted for in a way that guides the search?

3. Tool Appropriateness
What to assess: Whether the right sources are chosen for the job.
Look for: Use of authoritative, relevant resources (e.g., official filings for financial data, academic databases for research), with awareness of how to access them. Bonus if it considers alternatives when a source might fail.
Ask yourself: Are these the most trustworthy and efficient places to get this information?

4. Logical Ordering
What to assess: The flow of reasoning from start to finish.
Look for: A natural progression—starting broad or with high-signal clues, then narrowing down step by step. Each action should set up the next, not repeat or jump ahead.
Ask yourself: Does the sequence feel like a smart, efficient path to the answer?

5. Actionability
What to assess: Whether the plan can actually be carried out.
Look for: Concrete, unambiguous instructions that a person (or agent) could follow without guessing. Vague phrases like “look it up” weaken this dimension.
Ask yourself: Could someone execute this as written—or would they need to fill in gaps?

6. Clarity and Conciseness
What to assess: How easy the plan is to read and follow.
Look for: Clean structure, consistent language, and no unnecessary repetition. Good plans are brief but complete—nothing missing, nothing extra.
Ask yourself: Is this plan easy to understand at a glance?

Output Format:
Please output your evaluation in **JSON format** with the following structure:

{
  "Goal Alignment": {
    "score": [SCORE],
    "comment": "[COMMENT]"
  },
  "Subgoal Coverage": {
    "score": [SCORE],
    "comment": "[COMMENT]"
  },
  "Tool Appropriateness": {
    "score": [SCORE],
    "comment": "[COMMENT]"
  },
  "Logical Ordering": {
    "score": [SCORE],
    "comment": "[COMMENT]"
  },
  "Actionability": {
    "score": [SCORE],
    "comment": "[COMMENT]"
  },
  "Clarity and Conciseness": {
    "score": [SCORE],
    "comment": "[COMMENT]"
  },
  "total\_score": [TOTAL SCORE],
  "overall\_comment": "[OVERALL COMMENT]"
}

Where:

[SCORE] is a number from 0 to 5 based on the evaluation criteria.

[COMMENT] provides an explanation of the score.

[TOTAL SCORE] is the sum of the individual scores.

[OVERALL COMMENT] is a brief comment summarizing the overall quality of the plan.
\end{tcolorbox}

\subsection{Pseudo code of Plan rubrics optimization}

\begin{algorithm*}[htbp]
\caption{Stage 1: Anchor Plan Optimization via Rubric-Guided Learning}
\label{alg:stage1}
\begin{algorithmic}[1]
\Require Insight set $S = \{s_i = (q_i, p_i, \text{insight}_i)\}_{i=1}^n$ from prior trajectories 
\Require Initial rubrics $\mathcal{R}_0$ over planning dimensions $\{d_1, \dots, d_m\}$ 
\Require LLM-based updater $\mathcal{F}_{\text{Update}}$, convergence criterion

\State Initialize rubrics: $\mathcal{R} \gets \mathcal{R}_0$
\Repeat
    \State Sample balanced batch $\mathcal{B}_t = \{\mathcal{B}_{\text{success}}, \mathcal{B}_{\text{failure}}, \mathcal{B}_{\text{paired}}\} \subseteq S$
    \State Update rubrics: $\mathcal{R} \gets \mathcal{F}_{\text{Update}}(\mathcal{R}, \mathcal{B}_t)$
    \State Evaluate $\mathcal{R}$ on: \\
    \quad (i) alignment with human judgments \\
    \quad (ii) discriminative power between correct/incorrect plans
    \If{human feedback needed}
        \State Refine ambiguous/erroneous rubric items via annotators
    \EndIf
\Until{convergence criteria met}

\Ensure Final rubric set $\mathcal{R}$
\end{algorithmic}
\end{algorithm*}

\end{document}